\documentclass[twocolumn, switch]{article} 

\usepackage{amsmath, amsthm, amssymb, amsfonts}

\usepackage[numbers,square]{natbib}
\usepackage[utf8]{inputenc}	
\usepackage[T1]{fontenc}	
\usepackage{xcolor}		
\usepackage{hyperref}
\usepackage{multirow}
\usepackage{booktabs} 		
\usepackage{nicefrac}		
\usepackage{microtype}		
\usepackage{lineno}		
\usepackage{float}			
\usepackage{algorithm}
\usepackage{lipsum}		
\usepackage{tabularray}
\usepackage{amsmath,amssymb,amsfonts}
\usepackage[normalem]{ulem}
\usepackage{algpseudocode}
\usepackage{hyperref}

\usepackage{newfloat}
\DeclareFloatingEnvironment[name={Supplementary Figure}]{suppfigure}
\usepackage{sidecap}
\sidecaptionvpos{figure}{c}

\usepackage{titlesec}
\titlespacing\section{0pt}{12pt plus 3pt minus 3pt}{1pt plus 1pt minus 1pt}
\titlespacing\subsection{0pt}{10pt plus 3pt minus 3pt}{1pt plus 1pt minus 1pt}
\titlespacing\subsubsection{0pt}{8pt plus 3pt minus 3pt}{1pt plus 1pt minus 1pt}

\usepackage{tikz,xcolor,hyperref}

\definecolor{lime}{HTML}{A6CE39}
\DeclareRobustCommand{\orcidicon}{
	\begin{tikzpicture}
	\draw[lime, fill=lime] (0,0) 
	circle [radius=0.16] 
	node[white] {{\fontfamily{qag}\selectfont \tiny ID}};
	\draw[white, fill=white] (-0.0625,0.095) 
	circle [radius=0.007];
	\end{tikzpicture}
	\hspace{-2mm}
}
\foreach \x in {A, ..., Z}{\expandafter\xdef\csname orcid\x\endcsname{\noexpand\href{https://orcid.org/\csname orcidauthor\x\endcsname}
			{\noexpand\orcidicon}}
}

\title{ProMRVL-CAD: Proactive Dialogue System with Multi-Round Vision-Language Interactions for Computer-Aided Diagnosis }

\usepackage{authblk}

\author[1, \dag]{Xueshen Li}
\author[1, \dag]{Xinlong Hou}
\author[2, *]{Ziyi Huang}
\author[1, *]{Yu Gan}

\affil[1]{Department of Biomedical Engineering, Stevens Institute of Technology}
\affil[2]{Nokia Bell Labs}

\begin{document}

\twocolumn[ 
  
\maketitle

\begin{abstract}
Recent advancements in large language models (LLMs) have demonstrated extraordinary comprehension capabilities with remarkable breakthroughs on various vision-language tasks. However, the application of LLMs in generating reliable medical diagnostic reports remains in the early stages. Currently, medical LLMs typically feature a passive interaction model where doctors respond to patient queries with little or no involvement in analyzing medical images. In contrast, some ChatBots simply respond to predefined queries based on visual inputs, lacking interactive dialogue or consideration of medical history. As such, there is a gap between LLM-generated patient-ChatBot interactions and those occurring in actual patient-doctor consultations. To bridge this gap, we develop an LLM-based dialogue system, namely proactive multi-round vision-language interactions for computer-aided diagnosis (ProMRVL-CAD), to generate patient-friendly disease diagnostic reports. The proposed ProMRVL-CAD system allows \textit{proactive} dialogue to provide patients with constant and reliable medical access via an integration of \textit{knowledge graph} into a \textit{recommendation system}. Specifically, we devise two generators: a Proactive Question Generator (Pro-Q Gen) to generate proactive questions that guide the diagnostic procedure and a Multi-Vision Patient-Text Diagnostic Report Generator (MVP-DR Gen) to produce high-quality diagnostic reports. Evaluating two real-world publicly available datasets, MIMIC-CXR and IU-Xray, our model has better quality in generating medical reports. We further demonstrate the performance of ProMRVL achieves robust under the scenarios with low image quality. Moreover, we have created a synthetic medical dialogue dataset that simulates proactive diagnostic interactions between patients and doctors, serving as a valuable resource for training LLM.
\end{abstract}

\vspace{0.35cm}

] 
\footnotetext{$^\dag$These authors are contributing equally. $^*$Corresponding authors: ygan5@stevens.edu, ziyi.huang@nokia-bell-labs.com.}

Automatically generating patient-friendly diagnostic reports is crucial for mitigating clinical shortages and facilitating patient-doctor communication. Despite increasing interest, few studies have investigated building reliable interactive medical dialogue systems for automatic diagnostic report generation, especially with the consideration of both clinical visuals and medical history. Existing work on interactive medical services mainly targets question-answering (QA) tasks \cite{srivastava2020automatized,singhal2023towards,singhal2023large,chen2023meditron}, which only takes textual inputs but leaves visuals unsupported. However, medical images, such as radiography, are considered critical references for disease diagnosis due to their rich visual and textual features. Thus, these models could only serve as \textit{knowledge retrieval systems} to answer health-related questions, rather than \textit{medical dialogue systems} for diagnostic report generation. Another direction of studies is to directly integrate vision-language models into computer-aided systems to support visual inputs on disease diagnosis \cite{zhao2023chatcad+,zhou2023skingpt4,zhu2023minigpt4,chen-etal-2021-cross-modal}. Due to restricted scopes, these studies could only provide \textit{passive} responses to questions and \textit{minimal visual} interactions with the patients in which only a single visual input serves as the starting point of diagnosis. As such, these approaches merely rely on the evidence from medical visuals, while entirely ignoring the medical history and symptoms of the patients for disease diagnosis. This will lead to unreliable diagnostic results as the latter are considered critical clues for real-world clinical practices \cite{fukuzawa2024importance, sandler1980importance}. 

Similar to real-world clinical trials, a reliable medical diagnostic/dialogue system should mimic the procedure of proactive dialogue during regular patient-doctor interactions to collect medical history and symptom information for diagnostic references, rather than simply based on the medical visuals. Herein, performing proactive features to guide instructive conversations in computer-assisted diagnosis systems is necessary. Without effective queries, patients can hardly provide accurate and sufficient disease-associated descriptions, due to a dearth of domain expertise. The development of proactive features requires the model to find the logic and coherence behind the questions and responses, which is different from the conventional vision-language tasks such as visual question answering (VQA) and image-to-text generation \cite{antol2015vqa, wu2022medical, goyal2017making,tewel2022zerocap, ben2019vqa} that mainly targets the questions/descriptions related to the input visuals. This adds additional challenges to the development of proactive medical dialogue systems, as it seeks models with professional capability to understand context, perform interactions, and communicate logically.

So far, efforts to apply large language models (LLMs) \cite{openai2023gpt,hoffmann2022training,touvron2023llama,chowdhery2023palm,lee2024llmcxr} for automatic visual description generation have demonstrated promise in computer-aided diagnosis. However, the deployment of LLMs in proactive medical dialogue systems is still in its infancy. Existing work in \cite{wang2023chatcad, zhao2023chatcad+} developed the first medical dialogue system to provide medical advice on three imaging domains using large foundation models. Along with clinical concepts and doctors’ notes, Zhou et al \cite{zhou2023skingpt4} developed an interactive dermatology diagnostic system by fine-tuning Mini-GPT \cite{zhu2023minigpt4} using large skin disease images to generate skin disease reports. Lee et al \cite{lee2024llmcxr} proposed an instruction-tuning strategy to enhance LLMs' understanding of medical images by expanding their reasoning capability on input visuals. While powerful, these previous studies mainly chose a passive interaction scheme, in which the ChatBot could only passively respond to the questions proposed by the patient. Hence, these models fail to capture the medical history or disease-associated information from the patients and thus lead to unsatisfactory diagnostic reports.

As discussed above, existing solutions for medical diagnosis generation are less ideal, due to the lack of proactive patient-ChatBot interactions to collect disease-associated information. To bridge this gap, we propose an LLM-based dialogue system, namely proactive multi-round vision-language interactions for computer-aided diagnosis (ProMRVL-CAD), to reliably perform disease diagnosis with both visual evidence and patients' information. Guided by a knowledge graph-based recommendation system, ProMRVL-CAD supports multi-round interactions on a mixture of multi-view images and text inputs to enhance diagnosis performance, showing promise to outperform methods taking a single image for the entire diagnosis process. In Figure \ref{fig1_high_level}, we highlight the distinction between our proactive medical dialogue system and existing studies. Our major contributions are summarized as follows:

\begin{figure}[t]
\centering
\includegraphics[width=0.45\textwidth]{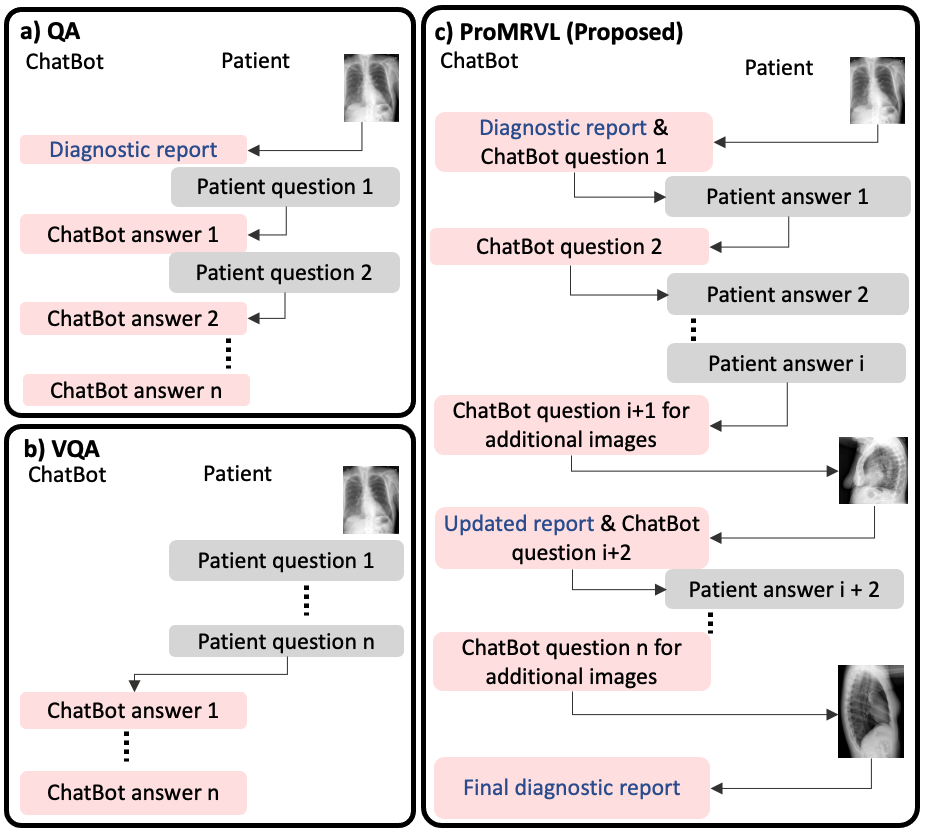}
\caption{High-level comparisons between existing works and ProMRVL-CAD when medical images are necessary for the diagnosis/treatment. \textbf{a):} Question-Answering (QA) task; \textbf{b):} Visual QA task; \textbf{c):} ProMRVL-CAD (Ours). By supporting multi-round text-visual inputs, ProMRVL-CAD mimics the consultation process between doctors and patients to collect diagnosis-orientated information. In contrast, existing frameworks simply use the first visual for report generation and then passively answer questions from the patients (QA) or only propose questions related to the input visuals (VQA). }
\label{fig1_high_level}
\end{figure}

\begin{itemize}
    \item We propose a novel Proactive Question Generator (Pro-Q Gen) to effectively collect critical inputs for disease diagnosis. The proactiveness of Pro-Q Gen stems from a knowledge graph built upon two real-world clinical datasets and a recommendation system to lead a productive medical information collection. This is significantly different from previous arts that can only perform diagnosis on input visuals. To our best, this study proposes the first proactive medical dialogue system for diagnostic report generation. 
    
    \item We propose a Multi-Vision Patient-Text Diagnostic Report Generator (MVP-DR Gen) that integrates multi-view visual and textual features without substantial modifications to existing LLMs. This integration of multi-view medical images and text enhances the effectiveness of multi-round proactive dialogues by mimicking a diagnostic process where a doctor considers the patient’s medical images, dialogue, and medical history. 

    \item Evaluating capability on two real-world publicly available datasets, MIMIC-CXR \cite{DBLP:journals/corr/abs-1901-07042} and IU-Xray \cite{DinaDemnerFushman.2015}, we validate the report quality and clinical efficiency on generated diagnostic reports. The experiments demonstrate the proactive interaction capacity of ProMRVL-CAD as well as its state-of-the-art performance on medical diagnostic report generation. We also validated the robustness of our framework on dataset with noisy and low resolution image resources.

    \item We develop a synthetic medical dialogue dataset, ProDial, containing multi-round conversations to mimic the proactive diagnostic interactions between patients and doctors in clinical practices. Based on the medical history and findings from the MIMIC-CXR dataset \cite{DBLP:journals/corr/abs-1901-07042}, the generated dialogue reflects the patients' information but carries complementary information to the medical visuals for disease diagnosis. This synthetic image dataset with dialogue is the first kind of this type in medical image analysis, addressing a need to pre-train large vision language model.   
    
\end{itemize}

\section{Methodology}
\subsection{System Overview}

\begin{figure*}[h]
\centering
\includegraphics[width=\textwidth]{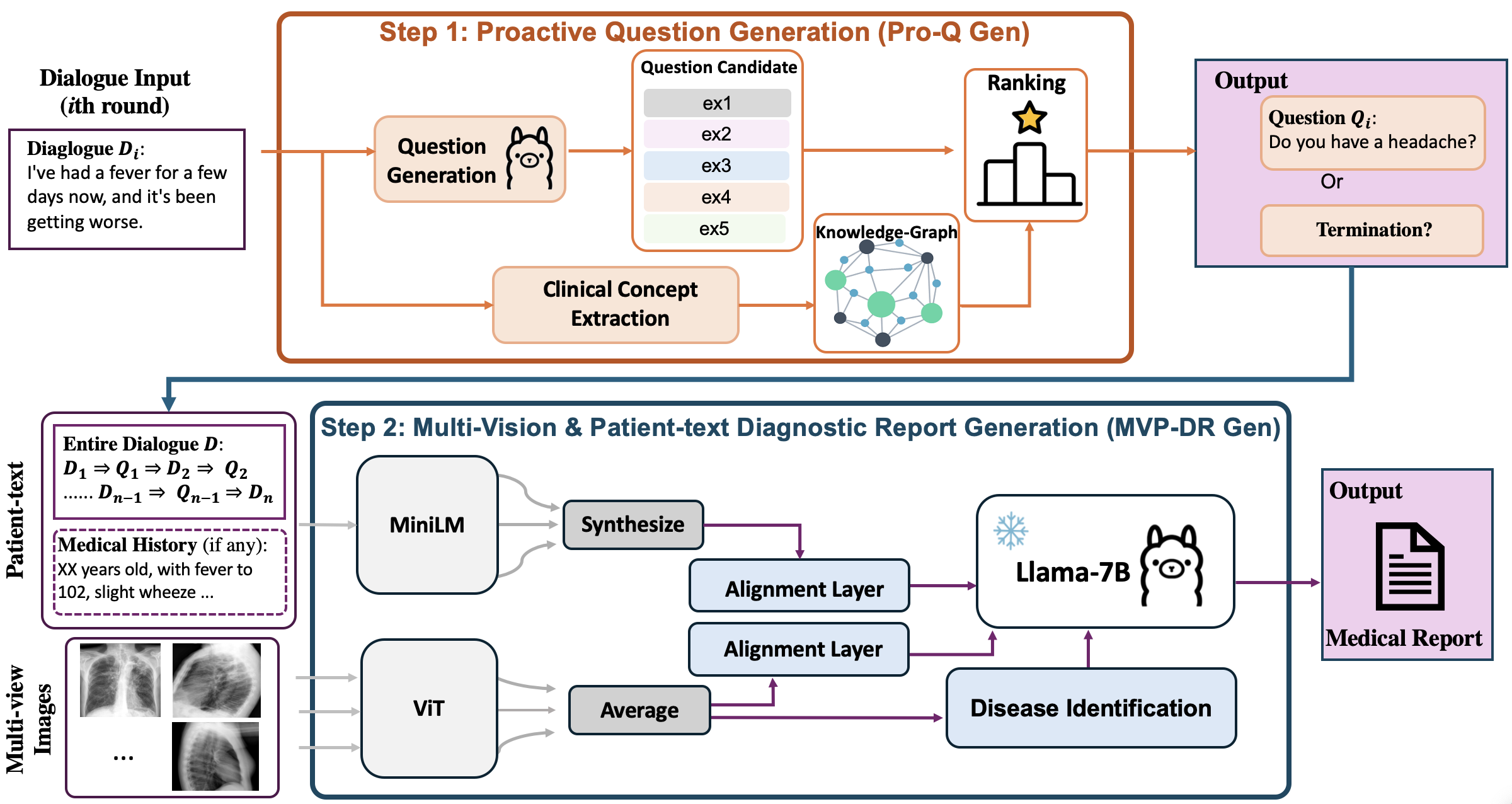}
\caption{The architecture of ProMRVL-CAD. It consists of two modules: Pro-Q Gen to prompt patients to provide more informative inputs and MVP-DR Gen to generate diagnostic reports from multi-modality inputs. }
\label{fig2_overall}
\end{figure*}

The overall structure of ProMRVL-CAD is illustrated in Figure \ref{fig2_overall}. The proposed system consists of two major components, Proactive Question Generator (Pro-Q Gen) for proactive conversation and Multi-Vision Patient-Text Diagnostic Report Generator (MVP-DR Gen) for disease diagnosis. For the Pro-Q Gen, we generate a synthetic and proactive medical dialogue dataset that is conditional on the medical history from a real-world medical dataset to mimic doctor-patient conversations. Then, we fine-tune a pre-trained LLM with the synthetic dialogue dataset and medical histories to boost its reasoning and interaction capability in disease analysis. In the MVP-DR Gen, we feed the medical visuals and the textual inputs (medical history and dialogue) to the second LLM. We confer the multi-modality capability on the model by leveraging a MiniLM \cite{10.5555/3495724.3496209} for dialogue analysis and deploying a vision transformer followed by an alignment layer for visual processing. Lastly, we freeze the parameters of the second LLM to fully utilize its language and reasoning capability. We train the rest of the components with the knowledge of the diagnostic reports to grant the proposed model a professional understanding of medical inputs.

\subsection{Step 1: Proactive Question Generation (Pro-Q Gen)}
\label{step1}
The design of ProMRVL-CAD is inspired by the conventional diagnostic procedure in which doctors inquire about the health conditions of patients. 
The goal of our proposed Pro-Q Gen is to proactively pose queries to acquire patient's health status information that underlying the potential disease. This requires Pro-Q Gen with strong disease reasoning capability to perform disease diagnosis, making it more challenging than classic recommendation tasks that mainly target feature embedding and pattern recognition. Different from existing work where the ChatBot passively answers queries, the Pro-Q Gen model generates questions from the ChatBot side to lead the conversation through a better understanding of patients' health conditions. In particular, we developed a dialogue recommendation system that generates proactive questions during conversations. Unlike a traditional end-to-end doctor agent, our system offers more precise query selection and refinement. As illustrated in Figure \ref{fig2_overall}, our system operates in two stages: generating query candidates and ranking them using a knowledge graph. This approach enables the proposed model to thoroughly explore the candidate query space generated by the foundation model and select the most relevant queries for the current response.

During the training process, we use cross-entropy to fine-tune the proposed LLMs to generate query candidates. The loss function is defined as 
\begin{equation}
L = \sum^L_{i=1} logP_\theta^{Q}(x_i; X_t^{Q}, X_{r,<i}^{Q}),   
\end{equation}
where $\theta^{Q}$ stands for trainable parameters, $x_i$ represents current predicted tokens with $i = 1, ..., L$ indicating the location of current token, $X_t^{Q}$ stands for the textual inputs, and $X_{r,<i}^{Q}$ represents the token before the predicted token.

\textbf{Question Generation.} We let the patient initiate the dialogue. Based on patients' input $D_i$, we generate $N$ queries as candidates. Then the patient agent generates another response for each dialogue $D_i$ in the queue. To generate valid question candidates, we fine-tuned the LLM using both the synthetic dialogue dataset and the medical history in the MIMIC-CXR dataset to enable it to generate proactive medical questions based on the responses of patients. To reduce the number of trainable parameters, we deploy the low-rank adaptation (LoRA) strategy \cite{Hu2021LoRALA, balazy2024lora} to inject trainable rank decomposition matrices into different layers. The use of medical history is inspired by the clinical practice in which doctors review patients' medical history before starting the intervention. As such, it could serve as a health condition constraint for the generation of proactive questions to improve the consistency between the dialogue and the patient's health condition. During the fine-tuning process, we use the responses from the patients in the synthetic dialogue dataset as the response and the questions from the doctor as the reference. Benefiting from our proactive dataset, our model could proactively raise questions to query patients' health conditions and ask for necessary medical visuals in multiple rounds. The termination of the conversation is determined by the confidence of the disease identification model on the collected evidence. 
\begin{figure}[t]
\centering
\includegraphics[width=0.45\textwidth]{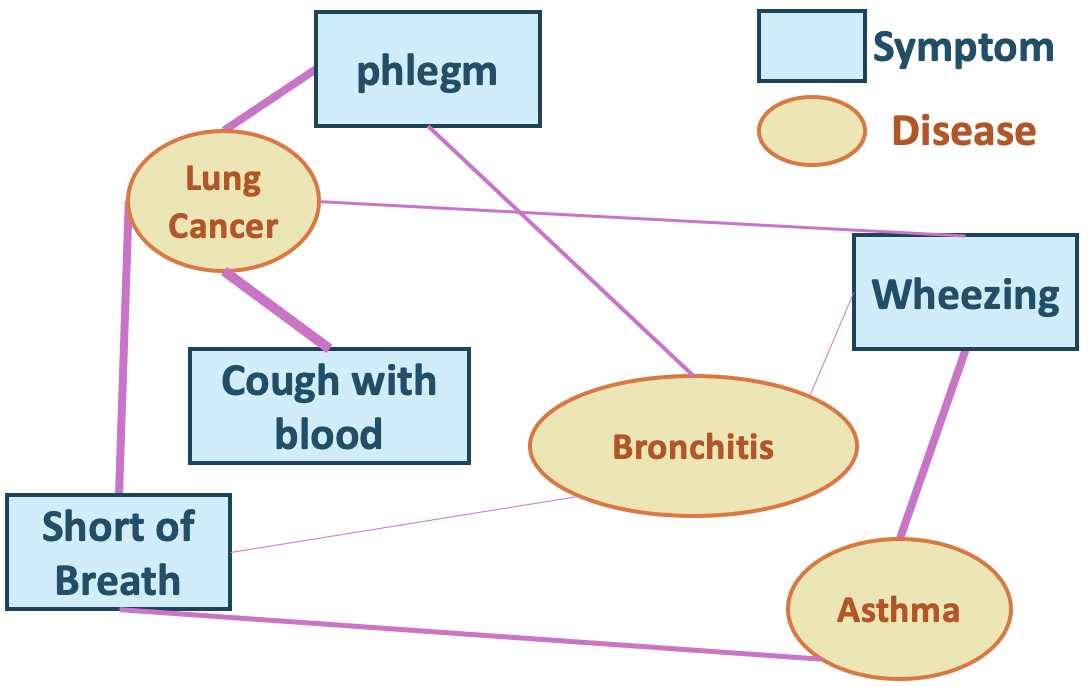}
\caption{An illustration of the knowledge graph with edge widths indicating the correlation between a disease and a symptom.}
\label{fig3_graph}
\end{figure}

\textbf{Candidate Ranking through Clinical Concept Knowledge Graph.} 
In our study, we propose to use knowledge graph \cite{Hogan_2021} to enhance the performance of our recommendation system. Inspired by \cite{zhang2016collaborative}, we develop a clinical concept knowledge graph to embed the structural knowledge between various diseases and symptoms. Note that our clinical concept graph is built upon an item graph, with no patient feature involved to protect users' privacy. The diseases and the corresponding symptoms are extracted from real-world clinical dialogue and medical records to ensure clinical professionalism. In Figure \ref{fig3_graph}, we show an example of our knowledge graph with edge widths indicating the correlation between the disease and the symptom. 

To fully utilize the language and reasoning capability of the pre-trained LLM and further reduce the retraining efforts, our clinical concept knowledge graph is integrated into the ranking stage to form a novel ranking criterion, rather than directly building into the candidate generation network for feature embedding. This is different from classic (item) knowledge graph-based recommendation systems that embed the item features through latent layers and generate recommendations through collaborative joint learning \cite{zhang2016collaborative, zhang2018learning, cao2019unifying, guo2020survey}. Our ranking criterion is based on the correlation between the potential diseases and the patient's symptoms. We first extract the patients' symptoms from past conversations and build a symptom base to record all symptoms from the patient. Then, we locate the most relevant disease through the knowledge graph and rank the candidate query through its correlation with the target disease. The candidate queries are ranked from 0 to 10 by a LLM, with 10 indicates the highest relevence and 0 indicates non-relevence. Specifically, our ranking system rejects repeat queries if the symptom has already been checked by the patient. Lastly, our system will terminate or move to the next potential disease if the most relevant symptoms have already been checked.

\begin{figure*}[t]
\centering
\includegraphics[width=0.9\textwidth]{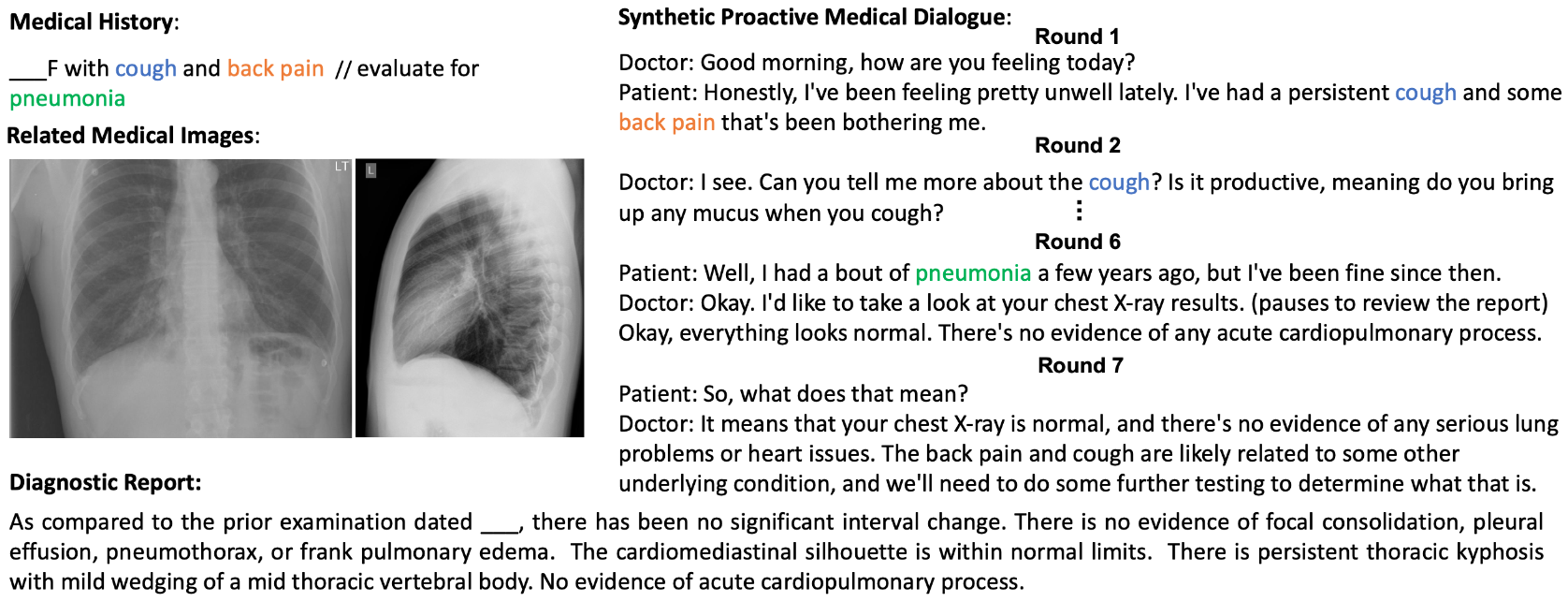}
\caption{Examples of synthetic proactive medical dialogue (training data) generated using medical history and medical images. The clinical concepts (highlighted by blue, orange, and green colors) are consistent among the medical history, the report, and the synthetic dialogue. Our synthetic dialogue is in line with the medical visuals and contains complementary information on health conditions to assist in disease diagnosis. }
\label{fig4_medialogue}
\end{figure*}

\textbf{Proactive Medical Dialogue Dataset Generation and Data Training on Hybrid Dataset.} Existing medical dialogue datasets only contain textual discussions of health conditions with limited descriptions of potential diseases \cite{zeng-etal-2020-meddialog,abacha2023empirical, yang2020generation, zhou2021generation}. In many cases, the conclusion of the discussion is just the initial guess of the potential disease, such as "Summary of the condition and initial impressions: Blepharitis" \cite{zeng-etal-2020-meddialog}. Moreover, these datasets do not include any visual evidence for diagnostic report generation and thus are not applicable to our study. As such, we generate a synthetic dataset to teach the model disease-related information and enable it to lead the conversation.  Inspired by the concept of VQA, we design an instructive protocol formulated by a virtual doctor and a patient to enable the model's reasoning and query capability to effectively collect the health condition from the patient. Based on the corresponding textual information, the contents of the dialogue are consistent with the medical visuals to assist further disease analysis. An example of synthetic proactive dialogue is shown in Figure \ref{fig4_medialogue}. Our Pro-Q Gen model is fine-tuned on a hybrid dataset that consists of both synthetic data and real-world medical conversation data \cite{zhang2023huatuogpt} to ensure its professionalism in medical dialogue generation.

\subsection{Step 2: Multi-Vision Patient-Text Diagnostic Report Generation (MVP-DR Gen)}\label{section_2_3}
The unique advantage of our proposed Multi-Vision Patient-Text Diagnostic Report Generator, namely MVP-DR Gen, lies in its multi-modality capability to simultaneously and synthetically process textual and visual inputs. Particularly, MVP-DR Gen could process medical visuals with different views if a single view is insufficient for disease diagnosis. As shown in Figure \ref{fig2_overall}, it employs a MiniLM \cite{10.5555/3495724.3496209, vergou2023readability} for textual analysis and a vision transformer (ViT) \cite{liu2021swin, yin2022vit, han2022survey} to learn the high-dimensional visual and texture features from the multi-view images. Inspired by \cite{covington2016deep}, the features extracted from different images/views are directly averaged to ensure a fixed input size for the followed vision-language generating task. Our vision-analysis task can be further divided into two sub-tasks, the disease identification task and the report generation task. Specifically, the report generation module aims to generate the diagnostic report from the embedded features, while the disease identification task contributes to improving the diagnosis capability. 
In the report generation task, we use an alignment layer, which serves as a soft prompt, to align the visual features with the LLMs. In this study, we use $Llama2-7B$ \cite{touvron2023llama, thakkar2023comprehensive} as our baseline LLM which has high efficiency to produce fast responses. During the entire training process, only the LLM remains frozen while the rest of the networks are randomly initialized. The loss function for the report generation task is jointly optimized by the combination of loss terms from each task. 

During the entire training process, only the LLM remains frozen while the rest of the networks are randomly initialized. The loss function for the report generation task is jointly optimized by the combination of loss terms from each task: 
\begin{equation}
\begin{aligned}[c]
L_{report}(\theta^{DR};X_r^{DR},X_v^{DR}, X_t^{DR},X_p^{DR})=&\\ 
    -\sum^{L}_{i=1}logp_{\theta^{DR}}(x_i;X_v^{DR},X_t^{DR},X_p^{DR},X_{r,<i}^{DR}) &, 
\end{aligned}
\end{equation}
where $\theta^{DR}$ represents the trainable parameters in the alignment layer, $L$ represents the length of the generated sentence, $X_r^{DR}$ represents the current prediction token, $X_v^{DR}$ represents the visual embedding, $X_t^{DR}$ represents the textual inputs, $X_p^{DR}$ represents prompts, and $X_{r,<i}^{DR}$ represents the token before the predicted token. We pose the disease identification task as a multi-label classification task to detect the potential disease in the input visuals. The model is optimized with the cross-entropy loss:
\begin{equation}
    L_{classification}(\hat{y}, y) = -\sum^C_{j=1}y_jlog(\hat{y}_j) + (1-y_j)log(1-\hat{y}_j),
\end{equation}
where $C$ stands for the number of classes, $\hat{y}$ denotes the prediction results, and $y$ denotes the ground truths. Overall, our loss function for the entire module is:
\begin{equation}\label{eq11}
    L = L_{classificaion} + \alpha L_{report}.
\end{equation}

\section{Experiments}

\begin{table*}[t!]
\caption{Baseline comparison on diagnostic report generation on MIMIC-CXR dataset. \textcolor{red}{Red} highlights the best performance and \textcolor{blue}{{blue}} highlights the second best performance.}
\label{ablation_comp}
\centering
\scalebox{1}{
\begin{tabular}{lccccc}
\hline
\multicolumn{1}{l}{Approaches} & BLEU1 $\uparrow$ & BLEU2 $\uparrow$ & BLEU3 $\uparrow$ & BLEU4 $\uparrow$ & ROUGE $\uparrow$ \\ \hline
R2GCMN (2021)                  & 0.353 & 0.218 & 0.148 & 0.106 & 0.278 \\
METrans (2023)                 & 0.386 & 0.250 & 0.169 & 0.124 & \textcolor{blue}{{0.291}} \\
ChatCAD+ (2023)                & 0.219 & 0.127 & 0.081 & 0.056 & 0.204 \\
R2GenGPT (2024)                & \textcolor{blue}{{0.405}} & \textcolor{blue}{{0.260}} & \textcolor{blue}{{0.178}} & \textcolor{blue}{{0.127}} & 0.290 \\
ProMRVL (Ours)                 & \textbf{\textcolor{red}{0.430}} & \textbf{\textcolor{red}{0.305}} & \textbf{\textcolor{red}{0.231}} & \textbf{\textcolor{red}{0.182}} & \textbf{\textcolor{red}{0.331}} \\ \hline
\end{tabular}
}
\end{table*}

\subsection{Experiment Setup}
\textbf{Datasets.} We carry out experiments using two publicly available datasets: MIMIC-CXR \cite{DBLP:journals/corr/abs-1901-07042} and IU-Xray \cite{DinaDemnerFushman.2015}.
We follow the same data partition policy in \cite{DBLP:journals/corr/abs-1901-07042} for the MIMIC-CXR dataset and the partition policy in \cite{Chen.2020} for the IU-Xray dataset. Additionally, we build two subset datasets, MIMIC-V2 and IU-V2, with cases that only contain frontal and lateral views of Xray images. 
Moreover, as mentioned in Section \ref{step1}, 
we generated a clinical dialogue dataset that contains both synthetic and real clinical dialogues to fine-tune our proposed Pro-Q Gen. Our clinical dialogue dataset consists of 78399 (66149 synthetic and 12250 real) clinical conversation records. For the synthetic dialogues, we used ChatGPT to generate dialogues using medical history from the MIMIC-CXR dataset. For the real clinical dialogues, we used part of the Huatuo-26M \cite{zhang2023huatuogpt} and CMtMedQA \cite{yang2024zhongjing} dataset, which is collected based on real conversations between the patient and doctor. 
Our diagnostic report generation task is evaluated on the MIMIC-CXR dataset and IU-Xray dataset, as they are among the largest real-world datasets consisting of Xray images and diagnosis reports. In the MIMIC-CXR dataset, each patient is associated with one or multiple Xray images, along with a diagnosis report containing impressions, findings, medical history, etc. The IU-Xray dataset consists of 7,470 Xray images, with 3,955 study cases and corresponding reports.

\textbf{Fine-tuning of Pro-Q Gen.} In the Pro-Q Gen module, we use Llama-3-8B Instruct \cite{llama3modelcard} as the backbone model for dialogue generation. During the fine-tuning process, we decompose the network parameter $\theta$ to $\theta_0+\Delta \theta(\Theta)$ using low-rank adaption with $N = 16$ ranks.

\textbf{Training of MVP-DR Gen.} We set $\alpha = 1$ for our overall loss function in Eq. \ref{eq11}. In the MVP-DR Gen module, we adopt the Swin Transformer \cite{9710580} pre-trained on ImageNet \cite{deng2009imagenet} as the ViT for feature embedding with a dimension of 1024. We deploy a linear layer for the alignment layer in the report generation and a fully connected layer as the classifier for disease identification. The disease identification results are transferred to the corresponding texts and integrated into the end of the generated report. We further deploy the MiniLM-L6-v2 \cite{10.5555/3495724.3496209} model to generate the word embeddings, with an output of 384 dimensions for each sentence. Similar to the visual task, the embeddings of sentences in the medical history and dialogues are averaged before sending to an alignment layer for downstream analysis. Lastly, the image embeddings and text embeddings are interrogated to a specific prompt to instruct Llama-2-7B for disease analysis and diagnostic report generation. The training process is conducted for 4 epochs for the MIMIC-CXR dataset, with a batch size of 6 and a learning rate of 1e-4. For the MIMIC-V2 and IU-V2 datasets, we fine-tune the models by using a batch size of 6 and a learning rate of 1e-4. During testing, we employ a beam search strategy with a beam size set to 3.

\textbf{Computational Resources.} The traing and testing experiments are carried out in parallel on 3 RTX A6000 GPUs (memory: 48GB) and 4 H100 Tensor Core GPUs (memory: 80GB). We use Intel(R) Core(TM) i9-10980XE CPU @ 3.00GHz with 36 cores, x86\_64 architecture. The operation system is Ubuntu 20.04.4 LTS.

\begin{table}[t]
\centering

\caption{Evaluation on clinical efficacy of the proposed model and baselines. \textcolor{red}{Red} highlights the best performance and \textcolor{blue}{{blue}} highlights the second best performance.}
\label{table_clinical1}
\scalebox{0.9}{
\begin{tabular}{llll} 
\hline
                         & F1  $\uparrow$  & Recall $\uparrow$ & Precision $\uparrow$  \\ 
\hline
R2GCMN, \textit{(2021)} & 0.334 & 0.275 & 0.278 \\
METrans, \textit{(2023)} & 0.364 & 0.309 & 0.311 \\
ChatCAD+, \textit{(2023)} & 0.290 & \textcolor{blue}{{0.403}} & \textcolor{blue}{{0.394}} \\
R2GenGPT, \textit{(2024)} & \textcolor{blue}{{0.392}} & 0.387 & 0.389 \\
ProMRVL (Ours) & \textcolor{red}{0.416} & \textcolor{red}{0.426}  & \textcolor{red}{0.406}      \\
\hline
                         &       &        &           
\end{tabular}
}
\end{table}

\textbf{Evaluation Metrics.} We measure the quality of the generated Dialogue dataset using quantitative scores provided by ChatGPT \cite{openai2023gpt}. Following the evaluation methodology in existing work \cite{yang2024zhongjing}, we scored 100 cases of dialogues between patients and doctors and compare the scores among our Dialogue dataset, MedDialog \cite{zeng-etal-2020-meddialog} and MTS-Dialog \cite{abacha2023empirical}.
For the diagnostic report evaluation, we measure the quality of the generated medical reports from two perspectives: report text quality and clinical efficacy. We adopt bilingual evaluation understudy (BLEU) \cite{papineni-etal-2002-bleu, reiter2018structured} and recall-oriented understudy for gisting evaluation (ROUGE) \cite{lin-2004-rouge, ng2015better} to evaluate the text quality of generated diagnostic reports and proactive questions. In addition to the text quality, we also use classification evaluation metrics, including precision, recall, and F1 scores, to evaluate the clinical efficacy of disease identification. We use CheXpert \cite{Irvin2019CheXpertAL, smit2020chexbert} for annotating the generated reports, which is compared to the ground truth annotations across 14 distinct categories (atelectasis, cardiomegaly, consolidation, etc.). 

\subsection{Evaluation on Diagnostic Report Generation} 
\textbf{Comparison with SOTAs.} We compare the diagnostic report generation performance of ProMRVL-CAD with the following SOTA: R2GCMN \cite{chen-etal-2021-cross-modal}, 
METransformer \cite{10203079}, ChatCAD+ \cite{zhao2023chatcad+}, and R2GenGPT \cite{ZhanyuWang.2023}. In Table \ref{ablation_comp}, we present the overall performance of ProMRVL-CAD and baselines on both report text quality. Our model achieves the best performance in all evaluation metrics regarding BLEU and ROUGE, indicating high text quality in comparison with original medical report. Moreover, we comapre the clinical efficacy results in Table \ref{table_clinical1}. We observed that the clinical efficacy of the proposed model is much higher than baselines. Particularly, our model shows a significant improvement in Recall. These results evidently demonstrate our effectiveness in disease diagnosis and abnormality detection, as Recall directly indicates the disease detection capability. 


\begin{table*}[t]
\caption{Ablation study on diagnostic report generation on MIMIC-CXR dataset. \textcolor{red}{Red} highlights the best performance and \textcolor{blue}{{blue}} highlights the second best performance.}
\label{ablation}
\centering
\centering
\scalebox{0.8}{\begin{tblr}{
  column{even} = {c},
  column{3} = {c},
  column{5} = {c},
  column{7} = {c},
  column{9} = {c},
  cell{5}{5} = {fg=blue},
  cell{5}{6} = {fg=blue},
  cell{5}{7} = {fg=blue},
  cell{5}{8} = {fg=blue},
  cell{5}{9} = {fg=blue},
  cell{8}{5} = {fg=red},
  cell{8}{6} = {fg=red},
  cell{8}{7} = {fg=red},
  cell{8}{8} = {fg=red},
  cell{8}{9} = {fg=red},
  hline{1-2,9} = {-}{},
}
Approaches      & Single-View & Multi-View & Textual Input & BLEU1 $\uparrow$ & BLEU2 $\uparrow$ & BLEU3 $\uparrow$ & BLEU4 $\uparrow$ & ROUGE $\uparrow$ \\
ChatCAD+ (2023) & \checkmark           &            &               & 0.219            & 0.127            & 0.081            & 0.056            & 0.204            \\
ChatCAD+ (2023) & \checkmark           &            & \checkmark             & 0.329            & 0.194            & 0.121            & 0.070            & 0.299            \\
R2GenGPT (2024) & \checkmark           &            &               & 0.405            & 0.260            & 0.178            & 0.127            & 0.290            \\
R2GenGPT (2024) & \checkmark           &            & \checkmark             & 0.416    & 0.270    & 0.187    & 0.135    & 0.299    \\
ProMRVL (Ours)  & \checkmark           &            &               & 0.361            & 0.228            & 0.156            & 0.112            & 0.264            \\
ProMRVL (Ours)  &             & \checkmark          &               & 0.370            & 0.238            & 0.166            & 0.123            & 0.273            \\
ProMRVL (Ours)  &             & \checkmark          & \checkmark             & \textbf{0.430}   & \textbf{0.305}   & \textbf{0.231}   & \textbf{0.182}   & \textbf{0.331}   
\end{tblr}}
\end{table*}

\textbf{Ablation Study.}
We conduct an ablation analysis to show the necessity of our modules and strategies. In Table \ref{ablation}, we present the performance of MVP-DR Gen with different model inputs. Compared with single-view images, we observe that using images with multiple views can potentially improve the diagnosis performance in the medical report generation. It is also noticed that textual input boosted the report generation even on the baseline models, which is also a demonstration of scalability of ProMRVL, indicating the feasibility of integrating the framework with other VLM. More importantly, these results confirm the necessity of using textual health status information for diagnosis report generation. To further demonstrate the impact of text input, we add two additional ablation studies on the latest two approaches, ChatCAD+ and R2GenGPT. It is worth noting that baselines of ChatCAD+ and R2GenGPT don't support text input themselves. We embedded the textual input in the same way of MVP-DR. We similarly noticed that text input could improve the performance of report generation. However, the improved performances in both ChatCAD+ and R2GenGPT are still lower than ProMRVL. This results indicated that although text input could improve the text quality, it it the unique multi-modal design in ProMRVL that boost the performance in report generation. Other methods (i.e., ChatCAD+ and R2GenGPT) could benefit from text input but not as much as ProMRVL. 

\begin{table}[t]
\centering
\caption{Quantitative evaluation on the synthetic dialogue dataset. The best performance is highlighted by \textcolor{red}{red}. The second best performance is highlighted by \textcolor{blue}{{blue}}.}
\label{datasetcompare}
\scalebox{0.8}{
\begin{tblr}{
  cell{1}{1} = {c},
  cell{1}{2} = {c},
  cell{1}{3} = {c},
  cell{2}{1} = {c},
  cell{2}{2} = {c},
  cell{2}{3} = {c},
  cell{3}{1} = {c},
  cell{3}{2} = {c,fg=blue},
  cell{3}{3} = {c, fg = blue},
  cell{4}{1} = {c},
  cell{4}{2} = {c,fg=red},
  cell{4}{3} = {c,fg=red},
  hline{1-2,5} = {1-3}{},
}
             & Professionalism  $\uparrow$   & Conciseness $\uparrow$    &  &  &  \\
MTS-Dialog (2023)         & 0.746           & 0.718          &  &  &  \\
MedDialog (2020) & {0.829}           & {0.788}          &  &  &  \\
ProDial (Ours)         & \textbf{0.875}  & \textbf{0.860} &  &  &  \\
             &                 &                &  &  &  
\end{tblr}
}
\end{table}
\subsection{Evaluation on Proactive Medical Dialogue}

\begin{figure*}[t]
\centering
\includegraphics[width=0.9\textwidth]{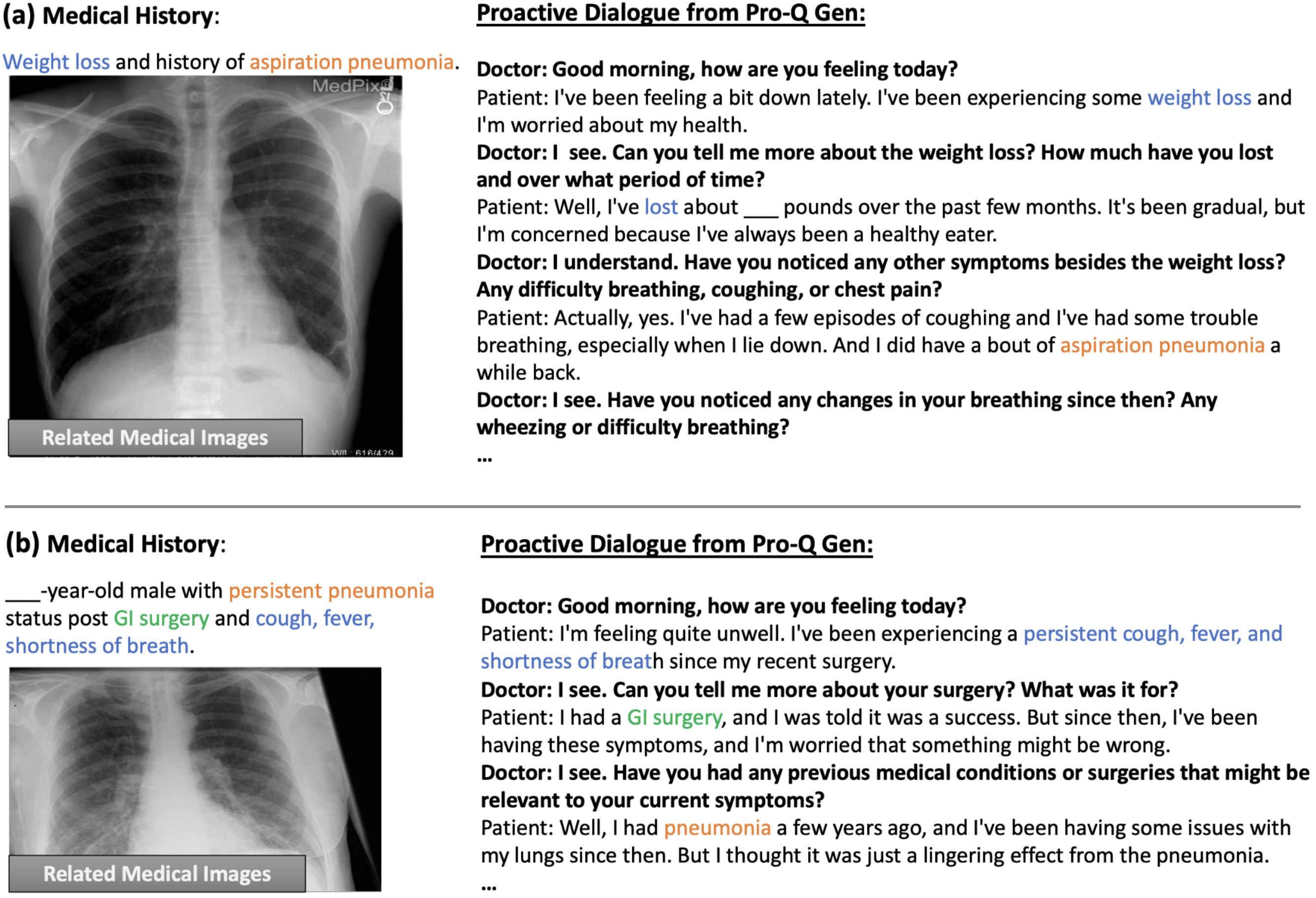}
\caption{Representative samples of the proactive dialogue produced by our proposed Pro-Q Gen. \textbf{(a).} A sample dialogue from a patient with weight loss and pneumonia. \textbf{(b).} A sample dialogue from a patient with several symptoms and various medical histories. Our proposed Pro-Q Gen could proactively pose queries to efficiently collect disease symptoms and medical history from the patients.}
\label{engage1}
\end{figure*}

\textbf{Quantitative Evaluation on Synthetic Medical Dialogue Dataset.} We generated 119,276 synthetic medical conversations based on corresponding files in the MIMIC-CXR dataset. In Table \ref{datasetcompare}, we show the quantitative evaluation of our synthetic medical dialogue dataset with two widely used medical dialogue datasets, MTS-Dialog \cite{abacha2023empirical} and MedDialog \cite{zeng-etal-2020-meddialog}.
We use Professionalism and Conciseness, which are considered key properties for natural and effective dialogue, to validate the quality of our synthetic dataset, ProDial. Same as \cite{zhang2023huatuogpt, yang2024zhongjing}, the Professionalism and Conciseness of the synthetic medical dialogue dataset are automatically evaluated by the quantitative scores provided by ChatGPT. As shown, our synthetic medical dialogue dataset has similar professionalism and conciseness properties as natural dialogue datasets. 
In Figure \ref{engage1}, we show two representative dialogue samples from our proposed Pro-Q Gen model, which provide immediate responses to the previous reactions.

\textbf{Quantitative Evaluation on Proactive Question Generator.} The query quality of the proposed Pro-Q Gen Model is evaluated on both language quality (Professionalism and Conciseness) and clinical professionalism (BLUEs and ROUGE) sides. In Table \ref{table4}, we compare the performance of the proactive questions generated by Pro-Q Gen with two state-of-the-arts, HuatuoGPT \cite{zhang2023huatuogpt} and Zhongjing \cite{yang2024zhongjing}. The Pro-Q Gen achieved higher performance on all evaluation metrics, showing its effectiveness in query generation. We further conduct an ablation study to show the necessity of our clinical concept knowledge graph on dialogue generation. As shown, introducing the clinical concept knowledge graph can significantly enhance clinical professionalism in query generation, with around $10\%$ performance improvements. 

\begin{table*}[t]
\centering

\caption{Baseline comparison and ablation study on proactive dialogue generation. The best performance is highlighted by \textcolor{red}{red}. The second best performance is highlighted by \textcolor{blue}{{blue}}.}
\label{table4}
\scalebox{0.8}{
\begin{tblr}{
  cell{2}{1} = {c},
  cell{2}{2} = {c},
  cell{2}{3} = {c},
  cell{2}{4} = {c},
  cell{2}{5} = {c},
  cell{2}{6} = {c},
  cell{2}{7} = {c},
  cell{2}{8} = {c},
  cell{3}{1} = {c},
  cell{3}{2} = {c},
  cell{3}{3} = {c},
  cell{3}{4} = {c},
  cell{3}{5} = {c},
  cell{3}{6} = {c},
  cell{3}{7} = {c},
  cell{3}{8} = {c},
  cell{4}{1} = {c},
  cell{4}{2} = {c},
  cell{4}{3} = {c},
  cell{4}{4} = {c},
  cell{4}{5} = {c},
  cell{4}{6} = {c},
  cell{4}{7} = {c},
  cell{4}{8} = {c},
  cell{5}{1} = {c,m},
  cell{5}{2} = {c,fg=red},
  cell{5}{3} = {c,fg=red},
  cell{5}{4} = {c,fg=blue},
  cell{5}{5} = {c,fg=blue},
  cell{5}{6} = {c,fg=blue},
  cell{5}{7} = {c,fg=blue},
  cell{5}{8} = {c,fg=blue},
  cell{6}{1} = {c},
  cell{6}{2} = {c,fg=blue},
  cell{6}{3} = {c,fg=blue},
  cell{6}{4} = {c,fg=red},
  cell{6}{5} = {c,fg=red},
  cell{6}{6} = {c,fg=red},
  cell{6}{7} = {c,fg=red},
  cell{6}{8} = {c,fg=red},
  hline{2-3,7} = {1-8}{},
}
                   &                                           &                                           &                         &                         &                         &                         &                         &  &  \\
                   & Professionalism $\uparrow$                           & Conciseness $\uparrow$                              & BLEU1 $\uparrow$                  & BLEU2 $\uparrow$                 & BLEU3 $\uparrow$                & BLEU4 $\uparrow$                  & ROUGE   $\uparrow$               &  &  \\
Huatuo (2023)            & 0.859                                     & 0.746                                     & 0.152                   & 0.066                   & 0.036                   & 0.023                   & 0.163                   &  &  \\
Zhongjing (2024)          & 0.641                                     & 0.682                                     & 0.115                   & 0.049                   & 0.024                   & 0.012                   & 0.119                   &  &  \\
{Pro-Q Model (Ours) \\ w/o knowledge graph} & \textbf{0.888} & \textbf{0.839} & {0.555} & {0.472} & {0.422} & {0.384} & {0.501} &  &  \\
Pro-Q Model (Ours)            & {0.880}                                    &  {0.826}                                  & \textbf{\textbf{0.622}}                            & \textbf{\textbf{0.532 }}                   & \textbf{\textbf{0.476}}                   & \textbf{\textbf{0.434}}                   & \textbf{\textbf{0.537}}                   &  &  \\
                   &                                           &                                           &                         &                         &                         &                         &                         &  &  \\
                   &                                           &                                           &                         &                         &                         &                         &                         &  &  
\end{tblr}}
\end{table*}

\subsection{Evaluation on Robustness of ProMRVL}

\textbf{Performance on a New Dataset.}
We conduct experiments on a similar Xray dataset, the IU-Xray dataset to show the robustness of our proposed ProMRVL-CAD system. The study cases with two views (frontal and lateral) images for the IU-Xray were selected and formed a new IU-V2 dataset.
Additionally, we create a sub-dataset of the MIMIC-CXR dataset, namely MIMIC-V2, which contains the study cases with two views.
In Table \ref{table2}, we show our evaluation results conducted on MIMIC-V2 and IU-V2 with different settings. 

\begin{table*}[t]
\centering
\caption{Baseline comparison on diagnostic report generation on two datasets with various experimental settings. \textcolor{red}{Red} highlights the best performance and \textcolor{blue}{{blue}} highlights the second best performance.}
\label{table2}
\scalebox{1}{
\begin{tblr}{
  column{3} = {c},
  column{4} = {c},
  column{5} = {c},
  column{6} = {c},
  cell{1}{2} = {c=2}{},
  cell{2}{1} = {r=5}{c},
  cell{2}{2} = {r=5}{c},
  cell{2}{6} = {fg=blue},
  cell{2}{7} = {fg=red},
  cell{3}{6} = {fg=blue},
  cell{3}{7} = {fg=red},
  cell{4}{6} = {fg=blue},
  cell{4}{7} = {fg=red},
  cell{5}{6} = {fg=blue},
  cell{5}{7} = {fg=red},
  cell{6}{6} = {fg=blue},
  cell{6}{7} = {fg=red},
  cell{7}{1} = {r=5}{c},
  cell{7}{2} = {r=5}{c},
  cell{7}{6} = {fg=blue},
  cell{7}{7} = {fg=red},
  cell{8}{6} = {fg=blue},
  cell{8}{7} = {fg=red},
  cell{9}{6} = {fg=blue},
  cell{9}{7} = {fg=red},
  cell{10}{5} = {fg=blue},
  cell{10}{7} = {fg=red},
  cell{11}{6} = {fg=blue},
  cell{11}{7} = {fg=red},
  hline{1-2,7,12} = {-}{},
}
              &                           &       & {ChatCAD+\\(2023)} & {R2GenGPT\\(2024)} & {MVP-DR\\(w/o Textual Input)} & {~ ~ MVP-DR\\(w/ Textual Input)} \\
{MIMIC\\~-V2} & {Report \\text \\quality} & BLEU1 $\uparrow$  & 0.228                     & 0.430                     & {0.438}            & ~ ~ ~ ~\textbf{0.481}          \\
              &                           & BLEU2 $\uparrow$  & 0.139                     & 0.290                     & {0.295}            & ~ ~ ~ ~\textbf{0.364}          \\
              &                           & BLEU3 $\uparrow$  & 0.094                     & 0.207                     & {0.212}                    & ~ ~ ~ ~\textbf{0.293}          \\
              &                           & BLEU4 $\uparrow$  & 0.068                     & 0.154                     & {0.159}                    & ~ ~ ~ ~\textbf{0.245}          \\
              &                           & ROUGE $\uparrow$  & 0.227                     & 0.318                     & {0.321}                    & ~ ~ ~ ~\textbf{0.392}          \\
IU-V2         & {Report \\text \\quality} & BLEU1 $\uparrow$  & 0.332                     & 0.454                     & {0.491}            & \textbf{~ ~ ~ ~0.522}          \\
              &                           & BLEU2 $\uparrow$ & 0.180                     & 0.294                     & {0.306}            & \textbf{~ ~ ~ ~0.375}          \\
              &                           & BLEU3 $\uparrow$ & 0.108                     & 0.211                     & {0.214}            & \textbf{~ ~ ~ ~0.293}          \\
              &                           & BLEU4 $\uparrow$ & 0.006                     & {0.160}             & 0.157                    & \textbf{~ ~ ~ ~0.242}          \\
              &                           & ROUGE $\uparrow$ & 0.250                     & 0.366                     & {0.375}            & \textbf{~ ~ ~ ~0.435}          
\end{tblr}}
\end{table*}

\begin{figure*}[t]
\centering
\includegraphics[width=1\textwidth]{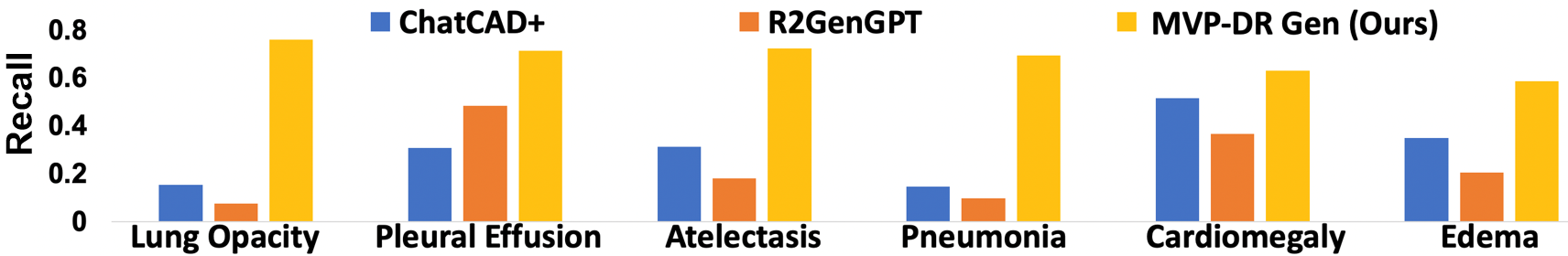}
\caption{Comparison of the clinical efficacy (recall) of the top-6 diseases in the MIMIC-V2 dataset.}
\label{compareRecall}
\end{figure*}
We confirm that the MVP-DR outperforms existing methods. Moreover, the textual input can further improve the performance of MVP-DR, using the same image input. Note that the text quality in Table \ref{table2} is higher than the multi-view results in Table \ref{ablation_comp}. This is because MIMIC-V2 is a subset of the MIMIC dataset where each data subject has two images. For comparison, the dataset used in Table \ref{table2} has a portion of subjects that only has one image to ensure a fair comparison with other approaches. The discrepancy between Table \ref{table2} and Table \ref{ablation_comp} highlights the need of building a multi-view dataset for diagnosis. Besides, the MVP-DR achieves much higher recall in detecting the top-6 diseases (lung opacity, pleural effusion, atelectasis, pneumonia, cardiomegaly, and edema) in the MIMIC-CXR dataset as shown in Figure \ref{compareRecall}, which takes up  $16.9\%$,  $14.8\%$,  $13.5\%$,  $13.3\%$,  $11.1\%$, and  $10.2\%$ of the positive cases in the MIMIC-V2 dataset. This indicates that our proposed MVP-DR Gen is less likely to generate misdiagnosed medical reports. Lastly, our model achieves similar high performance on both datasets, which evidently shows its generalization capability on diagnostic report generation.

\textbf{Scalability and Complexity}.
ProMRVL is generic and it can be easily scalable to other modules in terms of embedding, alignment, language model, etc. Streamlining the system's architecture could facilitate easier deployment and maintenance, enhancing its scalability. We notice that the integration of text input with a vision model in ProMRVL can also improve performance in ChatCAD+ and R2GenGPT, as we demonstrated in Table \ref{ablation}. To further improve the scalability, we conduct additional experiments to demonstrate the current network has room for a simplified implementation. We further reduce the complexity of the system using LoRA \cite{Hu2021LoRALA, balazy2024lora}
for Vision and LLM models. This approach reduces the parameters from \textbf{90.9M} to \textbf{5M}. The model performance, after reducing complexity nearly 20 times less, is satisfactory (\textbf{less than 4\%} overall performance drop) as shown in Table \ref{textual_model_scales}.

\begin{table*}[t]
\centering
\caption{Comparison the proposed model on different network scales.  \textcolor{red}{Red} highlights the best performance.}
\label{textual_model_scales}
\scalebox{1}{
\begin{tabular}{llllll}
\hline
                  & BLEU1          & BLUE2            & BLUE3           & BLEU4            & ROUGE          \\
\hline
Original (90.9 M) & \textcolor{red}{\textbf{0.430}} & \textcolor{red}{\textbf{0.305}}   & \textcolor{red}{\textbf{0.231}} & \textcolor{red}{\textbf{0.182}} & \textcolor{red}{\textbf{0.331}} \\
Simplified (5M)   & 0.413 & 0.295 & 0.230  & 0.189  & 0.323 \\
\hline             &                &                  &                 &                  & 
\end{tabular}
}
\end{table*}

\begin{table*}[t]
\caption{Evaluation on robustness of ProMRVL against image variation. \textcolor{red}{Red} highlights the best performance and \textcolor{blue}{{blue}} highlights the second best performance.}
\label{exp_noise_input}
\centering
\scalebox{1}{
\begin{tabular}{llllll}
\hline
            & BLEU1 & BLEU2 & BLEU3 & BLEU4 & ROUGE \\
\hline
Original dataset       & \textcolor{red}{\textbf{0.430}}           & \textcolor{red}{\textbf{0.305}}           & \textcolor{red}{\textbf{0.231}}           & \textcolor{red}{\textbf{0.182}}           & \textcolor{red}{\textbf{0.331}}           \\
Low resolution & 0.371           & 0.222           & 0.142           & 0.093           & 0.257           \\
Blurred        & \textcolor{blue}{{0.382}}           & \textcolor{blue}{{0.266}}           & \textcolor{blue}{{0.202}}           & \textcolor{blue}{{0.163}}           & \textcolor{blue}{{0.294}}           \\
\hline               &                &                &                &                &          
\end{tabular}
}
\end{table*}

\textbf{Robustness to Data Variability}. The generalization capability of our proposed method also lies in its robustness on data variability for different dataset. To test the robustness of ProMRVL, we conduct preliminary studies on two types of variations: 1) \textbf{low-resolution.} we tested the performance of ProMRVL on a dataset that the spatial resolution of is half of the original dataset; 2) \textbf{noisy images}. We examine the performance of ProMRVL on a dataset that the noise is added on original dataset. In this preliminary study, we set the noise as 7 photons variation at each pixel. The results are reported in In Table \ref{exp_noise_input}. Overall, the results show that our model is robust to the variations. For example, in low resolution case, BLEU1 is maintained to \textbf{86\%} of the performance on original dataset and in image with noisy quality, BLEU1 is maintained as \textbf{88.38\%} of the performance obtained from original dataset.

\section{Discussion}
In this study, we presented a novel ProMRVL model by introducing a proactive dialogue framework and multi-modal analysis, addressing significant challenges in medical dialogue and image integration. The framework's research paradigm and practical applications offer substantial advancements, such as improving interaction quality and mimicking real-world medical evaluations. These contributions highlight ProMRVL's capability to handle complex dialogic interactions and its potential for broader impact in medical AI.

Evaluation results validate the system's effectiveness and robustness. Comparative experiments reveal that ProMRVL outperforms baseline methods, even when enriched with textual input as shown in Table \ref{ablation}. Additionally, the inclusion of a knowledge graph significantly enhances diagnostic precision and recall metrics (i.e., BLEU and ROUGE), while maintaining professionalism and conciseness. Further experiments in Table \ref{exp_noise_input} and Table \ref{textual_model_scales} demonstrate the model's resilience to data variability, including low-resolution and noisy inputs, and its scalability, achieving satisfactory performance despite a significant reduction in system complexity using LoRA.

\section{Conclusion}
This paper devises a proactive dialogue system, ProMRVL-CAD, with the unique multi-modality feature of processing both clinical visuals and medical dialogue for disease diagnosis. The proposed model deploys a novel proactive question generator to mimic the nature of proactive dialogue during conventional patient-doctor interactions to collect the patients' health conditions. Our model has the superior performance of handling multi-image input over conventional LLMs with a nature of multi-round conversation. Evaluating on two publicly available datasets, we demonstrate that ProMRVL-CAD outperforms state-of-the-arts in generating medical reports. Our framework also creates the first synthetic proactive dialogue that integrates both medical visuals and patients' textual health status information from the existing clinical dataset. Future work will focus on implementing the system across multiple datasets empowered by federated learning. 

\bibliography{mybibliography}{}

\end{document}